\pdfoutput=1

\documentclass[11pt]{article}

\usepackage{acl2023}

\usepackage{times}
\usepackage{latexsym}

\usepackage[T1]{fontenc}

\usepackage[utf8]{inputenc}

\usepackage{microtype}

\usepackage{graphicx}
\usepackage{hyperref}
\usepackage{url}
\usepackage{algorithm}
\usepackage[noend]{algpseudocode}
\usepackage{multirow}
\usepackage{booktabs}
\usepackage{tablefootnote}
\usepackage{amsthm}
\usepackage[singlelinecheck=off]{caption}
\usepackage{amsmath}
\usepackage{subcaption}
\usepackage{enumitem}
\usepackage{float}
\usepackage{dblfloatfix}
\usepackage{placeins}
\usepackage[normalem]{ulem}

\def\*#1{\mathrm{\mathbf{#1}}}
\newcommand{\furl}[1]{\footnote{\url{https://#1}}}

\title{Finding the Pillars of Strength for Multi-Head Attention}

\author{Jinjie Ni \and
  Rui Mao \and
  Zonglin Yang \and
  Han Lei \and
  Erik Cambria \\
  Nanyang Technological University, Singapore\\
  \texttt{\{jinjie001, leih0003\}@e.ntu.edu.sg,}\\\texttt{\{rui.mao, zonglin.yang, cambria\}@ntu.edu.sg}}

\definecolor{bblue}{rgb}{0.54, 0.81, 0.94}

\definecolor{YELLOW}{rgb}{0.83, 0.61, 0.18}

\DeclareMathOperator*{\argmin}{arg\,min}

\makeatletter
\newcommand\blfootnote[1]{%
  \begingroup
  \renewcommand\thefootnote{}\footnote{#1}%
  \addtocounter{footnote}{-1}%
  \endgroup
}
\makeatother

\begin{document}

\maketitle



\begin{abstract}
Recent studies have revealed some issues of Multi-Head Attention (MHA), e.g., redundancy and over-parameterization. Specifically, the heads of MHA were originally designed to attend to information from different representation subspaces, whereas prior studies found that some attention heads likely learn similar features and can be pruned without harming performance. Inspired by the minimum-redundancy feature selection, we assume that focusing on the most representative and distinctive features with minimum resources can mitigate the above issues and lead to more effective and efficient MHAs. In particular, we propose Grouped Head Attention, trained with a self-supervised group constraint that group attention heads,
where each group focuses on an essential but distinctive feature subset. We additionally propose a Voting-to-Stay procedure to remove redundant heads, thus achieving a transformer with lighter weights. 
Moreover, our method achieves significant performance gains on three well-established tasks while considerably compressing parameters. The code\furl{github.com/Psycoy/ACL-2023-Grouped-Head-Attention} is released. \blfootnote{\textit{Proceedings of the 61st Annual Meeting of the Association for Computational Linguistics. Volume 1: Long Papers, pages 14526–14540.}}
\end{abstract}


\section{Introduction}
\label{sec: Introduction}
Transformers have shown promising performance across various tasks
. 
However, they have some issues, e.g., redundancy and over-parameterization, which is mainly caused by Multi-Head Attention (MHA)~\citep{DBLP:conf/nips/MichelLN19,DBLP:conf/acl/VoitaTMST19} and Feed-Forward Network (FFN)~\citep{DBLP:journals/corr/abs-1907-01470, DBLP:conf/iclr/WuFBDA19, DBLP:conf/iclr/WuLLLH20} of transformer. We aim to mitigate the redundancy and over-parameterization issues by optimizing the MHA module.
The multi-heads were originally designed to attend to different representation subspaces of input~\citep{DBLP:conf/nips/VaswaniSPUJGKP17}. However, prior works~\citep{DBLP:conf/nips/MichelLN19,DBLP:conf/acl/VoitaTMST19} have shown that the attention heads are highly redundant and over-parameterized after training because some heads can be switched off with a negligible performance drop.

Such an issue is probably caused by their parallel design: the heads naturally work in the same way and likely attend to similar features~\citep{DBLP:journals/corr/abs-2006-16362}.
The existing redundancy optimization methods are mainly based on homogenization, diversification, and head significance.
However, they all have some limits.
\textbf{(1)} The homogenization-based methods mitigate redundancy and over-parameterization by making heads similar and removing unnecessary parameters. \citet{DBLP:journals/corr/abs-2006-16362} homogenized attention heads by sharing most weights between all heads, which reduced the redundant parameters but sacrificed the performance somewhat because of the lack of diversity. \textbf{(2)} The diversification-based methods diversify the heads to enrich features and reduce the inter-head redundancy. \citet{DBLP:conf/emnlp/LiTYLZ18} found that diversifying attention heads by adding a regularization could force MHA to reduce inter-head information redundancy, yielding performance gains in Machine Translation. However, such strategy that retains all feature subsets is sub-optimal, because it does not address the issue that MHA is over-parameterized.
\textbf{(3)} The significance-based methods~\citep{DBLP:conf/nips/MichelLN19, DBLP:conf/acl/VoitaTMST19, DBLP:journals/tacl/LiCS21} learn significance scores for the heads to prune unimportant ones.
However, the retained important heads still remain inter-head redundancy without diversifying them.


Considering the issues of the above-mentioned methods, we hypothesize that attending to the most representative and distinctive feature subsets with minimum resources leads to more effective and efficient MHAs, which is inspired by the minimum-redundancy feature selection~\citep{DBLP:journals/corr/abs-2006-16362}. Accordingly, we propose a divide-and-conquer strategy, including Group-Constrained Training (GCT) and Voting-to-Stay (V2S), to achieve the setting of our assumption and mitigate the above-mentioned issues.
We illustrate them below.

We first propose a strategy to group and distinguish attention heads, where a Grouped Head Attention (GHA) is obtained via the self-supervised GCT. By encouraging homogenization within a group and diversification between groups, the MHA is forced to divide its heads to work in several separate groups, where each group focuses on an essential but unique feature subset, being in line with the setting of our assumption.
Note that the redundancy exists when the resources deployed by the model are more than enough to process current information 
\cite{DBLP:journals/corr/abs-2006-16362}. 
GHA reduces the redundancy in two aspects:
\begin{itemize}[leftmargin=*]
\item The intra-group homogenization reduces redundancy by encouraging similar intra-group heads and only retaining the most representative one later to lower the resource deployment. 
\item The inter-group diversification reduces redundancy by forcing heads to attend to more diversified features (with less overlap between heads) so that the unique information to process increases and matches the resources deployed. 
\end{itemize}
Next, we show that GHA-PS (GHA with the Pillar of Strength), a lighter-weight GHA, can be achieved by excluding the redundant heads of GHA via the V2S procedure.
V2S culls the redundant heads that share similar patterns with the most representative head (PS head) of a group, which is selected by voting on different training batches.
Note that upon the convergence of the GCT, the heads are highly homogenized within a group, thus being redundant because they process similar information. As a result, once the redundant heads are culled, the PS heads can still achieve the essential utility of the original attention layer and yield comparable performance to the unculled model. The Lottery Ticket hypothesis~\citep{DBLP:conf/iclr/FrankleC19} argues that subnetworks in an over-parameterized neural network can converge faster and achieve comparable or better performance than the original network. Our GHA-PS achieving better results is also in line with this hypothesis.

Such a divide-and-conquer combination resolves the issues of previous redundancy optimization methods:
\textbf{(1)} Our model achieves better parameter efficiency, resolving the issue of previous diversification-based methods; \textbf{(2)} The feature diversity is guaranteed and the inter-head redundancy is reduced, resolving the problems of previous homogenization- and significance-based methods.

We evaluate our method on three benchmarking tasks. We denote the corresponding transformer architectures of GHA and GHA-PS as Grouped Head Transformers (GHT) and Grouped Head Transformers with the Pillars of Strength (GHT-PS), respectively.
GHT and GHT-PS achieve significant improvements over the strong baselines in Machine Translation (MT) BLEU scores (+3.8\% and +4.4\% averaged on 7 datasets), Language Modeling (LM) perplexity (-2.8\% and -2.9\%), and Abstractive summarization (AS) F1-Rouge (+6.7\% and +7.0\% on average). GHT-PS exhibits higher efficiency in model size, inference speed, and floating-point operations (FLOPs). The light architecture of GHT-PS reduces 63.6\% parameters of the vanilla transformer and yields comparable performance.
The key contributions of our work are threefold: 

\begin{itemize}[leftmargin=*]
   \item We find that, in a certain range, higher compactness of attention heads (i.e., the intra-group heads become closer to each other and the inter-group ones become farther) improves MHA's performance, forcing MHA to focus on the most representative and distinctive features. It provides guidance for future architectural designs of MHA.
   \item We propose a divide-and-conquer strategy that consists of GCT and V2S. It mitigates the redundancy and over-parameterization issues of MHA. Our method uses fewer parameters and achieves better performance, outperforming the existing MHA redundancy/parameter reduction methods. 
   \item We verify our methods on three well-established NLP tasks. The superior results on datasets with multiple languages, domains, and data sizes demonstrate the effectiveness of our method.
\end{itemize} 


\section{Related Work}
\label{Related Work}

\paragraph{Parameter efficiency.} 
Different methods were proposed to achieve lightweight transformers: \textbf{(1)} replacing attention with lightweight modules, e.g., convolution modules, such as Dynamic Conv~\citep{DBLP:conf/iclr/WuFBDA19} and Lite Transformer~\citep{DBLP:conf/iclr/WuLLLH20}; \textbf{(2)} removing or replacing the feed-forward layers, such as~\citet{DBLP:journals/corr/abs-1907-01470} and~\citet{DBLP:conf/iclr/WuLLLH20}; \textbf{(3)} pruning the model, such as~\citet{DBLP:conf/nips/MichelLN19},~\citet{DBLP:conf/acl/VoitaTMST19}, and~\citet{DBLP:journals/tacl/LiCS21}.

\paragraph{Modified multi-head mechanism.}
\citet{DBLP:journals/corr/abs-1711-02132} learned to weight the projected output of different heads, performing weighted sum over them. \citet{DBLP:conf/naacl/LiYDWLT19} aggregated the output of different heads by dynamic routing;~\citet{DBLP:conf/emnlp/CuiIHUN19} used different attention mechanisms, e.g., global/local and forward/backward attention for different heads;~\citet{DBLP:journals/corr/abs-2003-02436} mixed different heads before and after the softmax operation in an attention function to achieve communication between heads.

\paragraph{Head redundancy optimization.}~\citet{DBLP:conf/nips/MichelLN19} and~\citet{DBLP:conf/acl/VoitaTMST19} found that only a subset of the attention heads have significant utilities in transformer, where the important heads could be identified by Expected Sensitivity and Layer-wise Relevance Propagation (LRP)~\citep{ding-etal-2017-visualizing}. Upon this,~\citet{DBLP:journals/tacl/LiCS21} learned per-head importance scores and pruned the heads.
\citet{DBLP:journals/corr/abs-2006-16362} homogenized the attention heads by sharing a part of the weights between heads, which lowered the number of parameters but sacrificed performance. \citet{DBLP:conf/emnlp/LiTYLZ18} found that diversifying attention heads by adding a regularization can force MHA to reduce inter-head redundancy, yielding performance gains for Machine Translation. 
However, previous methods either traded performance for efficiency or retained extra parameters.


\section{Methodology}

\begin{figure}[t]
\begin{center}
\includegraphics[width=0.46\textwidth]{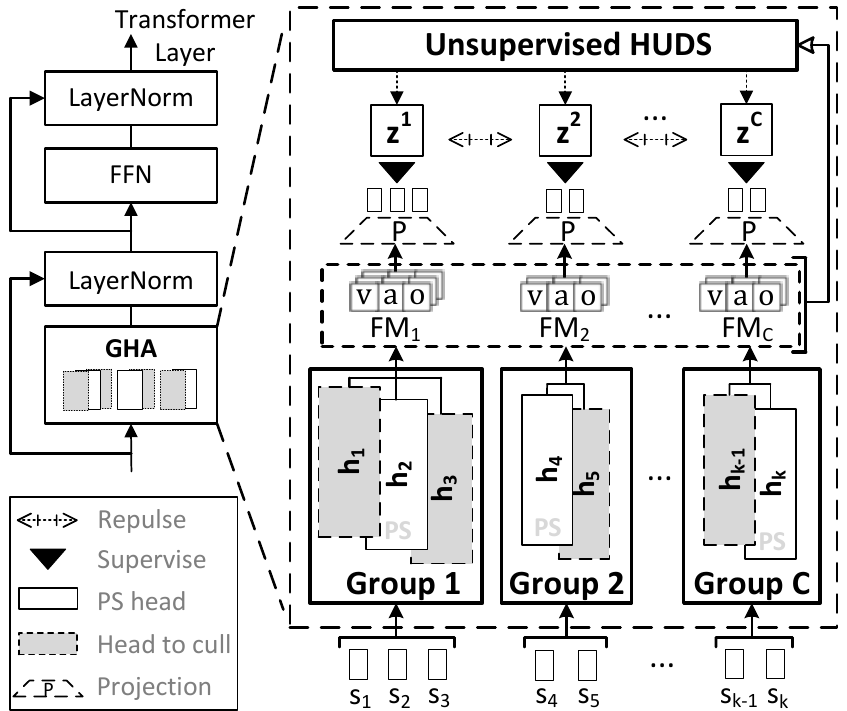}
\caption{The Grouped Head Attention. The heads in a group are under self-supervision of the discovered group hidden units $\*Z$ (Eq.\ref{eq: grouping}). The non-PS heads (dashed gray boxes in a group) will be culled in the VS procedure (Algorithm~\ref{Voting to Stay}). \(S_k\) denotes the \(k\)-th representation subspace; FM\(_C\) denotes the \(C\)-th feature map group. The output of GHA is omitted for simplicity.}
\label{figModelArch}
\end{center}
\end{figure}

There are two core components in our method, namely the Group-Constrained Training (GCT) and the Voting-to-Stay (V2S) procedure. GHA (Figure~\ref{figModelArch}) is developed with GCT that removes head redundancy; GHA-PS is developed by removing the redundant parameters of GHA in V2S. In this section, we detail the process of developing the GHA and finding its Pillars of Strength (PS).

\subsection{Grouped Head Attention with Hidden Units}
\label{sec:GHA}

First, we detail the core module of GHT, the GHA with hidden units, which is built based on MHA via the GCT. The GCT divides the attention heads of MHA into several groups and makes heads within a group become more similar, whereas heads between groups become more different. Thus, MHA is forced to divide its heads to work in several separate groups, where each group focuses on an essential but unique feature subset to reduce head redundancy. We will show the effectiveness in \S~\ref{Results}. 

Given a transformer model \(f(\*x; \*\theta)\) with 
\(n\) attention layers,
the set of heads at attention layer \(l\) is denoted as \(\*H_l\) \(= \{\*h_{1, l} ,..., \*h_{k, l}\}\), where $k$ is the number of heads. The outputs of the attention heads are concatenated and projected with $W^{out}$, where the \(i\)-th head output \(\*o_{i,l}\) in layer $l$ results from the computation of the projection matrices \(W^\*Q_{i, l}\), \(W^\*K_{i, l}\), and \(W^\*V_{i, l}\) of this head: 
\begin{equation}
    M\!H\!A_{l}(\*Q, \*K, \*V) = Concate(\*o_{1, l}, ..., \*o_{k, l})W^{out}
\end{equation}
\begin{equation}
    \*o_{i, l}= softmax(\frac{(\*QW^\*Q_{i,l})(\*KW^\*K_{i,l})^T}{\sqrt{d_k}})(\*VW^\*V_{i,l}).
\end{equation}
Three feature maps (FMs) of GHA are extracted for the self-supervised GCT: \textbf{(1)} the result of \(\*VW^\*V_{l}\), denoted as \(\hat{\*V}_l\) = \(\{\*v_{1, l}, ..., \*v_{k, l}\}\) (the value FM); \textbf{(2)} the attention weights of the \(l\)-th layer, denoted as \(\*A_l \)= \(\{\*a_{1, l}, ... , \*a_{k, l}\}\) (the attention FM); \textbf{(3)} the output of the \(l\)-th layer before the output projection $W^{out}$, denoted as \(\*O_l \)= \(\{\*o_{1, l}, ... , \*o_{k, l}\}\) (the head output FM). Moreover, \(\hat{\*V}\) = \(\{\hat{\*V}_1, ... , \hat{\*V}_l\}\), \(\*A\) = \(\{\*A_1, ... , \*A_l\}\), \(\*O\) = \(\{\*O_1, ... , \*O_l\}\). 
Given the FMs, a Hidden Unit Discovery System (HUDS) \(\Omega\) assigns a hidden unit \(\*z^j_{i, l}\) for each head to represent its group property, where $i$ denotes the $i$-th head and $j$ denotes the $j$-th group hidden unit. $\*z^j_{i, l} \in \hat{\*Z}_l$, where $\hat{\*Z}_l = \{\*z^1_{l}, ... , \*z^C_{l}\}$ represents the hidden unit candidates, and the hidden units assigned to the heads are denoted as \(\*Z_l = \{\*z_{i, l}, ... , \*z_{i, l}\}\). 

$\*Z_l$ is discovered by the HUDS $\Omega$: \(\*Z_l = \Omega(\*E_{l})\), where \(\*E_{l}\)
denotes either one of the \(\hat{\*V}_l\), \(\*A_l\), or \(\*O_l\). Here $\Omega(\cdot)$ is an unsupervised algorithm that divides the heads into $C$ groups given their FMs, such as K-means\footnote{K-means fixes the group numbers for fair comparisons in \S\ref{Results}. Other clustering algorithms may also be applicable.}:
\begin{equation}
    \Omega(\*E_{l}) = \argmin_{\*Z_l} \sum_{i=1}^C \sum_{\*x \in \hat{\*E}^i_{l}} ||\*x-\mu_i||^2,
\end{equation}
where $\hat{\*E}^i_{l}$ is the set of feature maps of the $i$-th head group in the $l$-th attention layer. Then, the feature map groups of the $l$-th attention layer are denoted as $\hat{\*E}_l = \{\hat{\*E}^1_{l}, ..., \hat{\*E}^i_{l} , ..., \hat{\*E}^C_{l}\}$.  $\mu_i$ is the mean of the feature map vectors in $\hat{\*E}^i_{l}$.
 The hidden units $\*Z = \{\*Z_1, ..., \*Z_l\}$ are \(C\)-class categorical variables (Eq.\ref{eq: grouping}(A)) or continuous vectors (Eq.\ref{eq: grouping}(B)) to supervise the GCT. The objective of the self-supervised GCT is termed as:
\begin{align}
\small
\label{eq: grouping}
\begin{split}
		\displaystyle &L_z(\*f; \*A, \hat{\*V}, \*O, \*Z) =\\
                &\begin{cases}
                   \displaystyle -\frac{1}{kn} \alpha \sum_{l=1}^n \sum_{i=1}^k log\ p_z(\*z_{i, l} | \*v_{i, l}, \*a_{i, l}, \*o_{i, l})\ +\\
                   \displaystyle \frac{1}{(C-1)kn} \beta \sum_{l=1}^n \sum_{i=1}^k \sum_{j_2\neq j_1} log\ p_z(\*z^{j_2}_{l} | \*v^{j_1}_{i, l}, \*a^{j_1}_{i, l}, \*o^{j_1}_{i, l}) & \text{(A)}\\ 
                    \displaystyle \quad\! \frac{1}{kn} \alpha \sum_{l=1}^n \sum_{i=1}^k \varphi(\*v_{i, l}, \*a_{i, l}, \*o_{i, l}; \*z_{i, l})\ -\\
                    \displaystyle  \frac{1}{{C \choose 2} n} \beta \sum_{l=1}^n \sum_{j_1=1}^{C-1} \sum_{j_2=j_1+1}^C \varphi(\*z^{j_1}_{l}; \*z^{j_2}_{l})
                    & \text{(B)}
                \end{cases}
\end{split}
\end{align}
Either when $\*Z$ are categorical variables (Eq.\ref{eq: grouping}(A)) or continuous vectors (Eq.\ref{eq: grouping}(B)), the objective is composed of a homogenization term and a diversification term\footnote{The coefficients \(\alpha\) and \(\beta\) of Eq.\ref{eq: grouping} respectively control the intra-group homology and inter-group diversity degrees to achieve different group intensities in different tasks/datasets.}. 
$\*v^{j}_{i, l}$, $\*a^{j}_{i, l}$, and $\*o^{j}_{i, l}$ denote the feature maps of the $i$-th head belonging to the \(j\)-th group. 
\(p_z(\*z_{i, l} | \*v_{i, l}, \*a_{i, l}, \*o_{i, l})\) denotes the predicted probability of the assigned group hidden variable \(\*z_{i, l}\), given \(\*v_{i, l}\), \(\*a_{i, l}\), and \(\*o_{i, l}\). \(\varphi(\*x; \*y)\) denotes a cosine similarity measurement between \(\*x\) and \(\*y\) (following~\citet{DBLP:conf/emnlp/LiTYLZ18}). \(\varphi(\*v_{i, l}, \*a_{i, l}, \*o_{i, l}; \*z_{i, l})\) \(=\) \(\tau_1 \varphi(\*v_{i, l}\); \(\*z_{i, l})\) \(+\) \(\tau_2 \varphi(\*a_{i, l}\); \(\*z_{i, l})\) \(+\) \(\tau_3 \varphi(\*o_{i, l}\); \(\*z_{i, l})\), where \(\tau\) is a coefficient, determined by the specific settings for each dataset \& task. When \(\*Z\) are categorical variables, the grouping is a classification task whose classification heads project the output into \(C\) classes; 
when \(\*Z\) are continuous vectors, the grouping process is a metric learning task 
whose similarity computations are conducted between \(\*Z\) and the projected FM representations. In both conditions, GHA is supervised by \(\*Z\) to make the heads in the same group yield similar patterns, whereas those in different groups repulse from each other. The overall objective is given by \(L = L_t +L_z\), where \(L_t\) is the task-specific objective.

\subsection{The Pillars of Strength}
Being consistent with Lottery Ticket hypothesis~\citep{DBLP:conf/iclr/FrankleC19}, we establish the GHT-PS from GHT as its subnetwork by removing redundant heads from GHA to achieve higher parameter efficiency. We propose the V2S procedure to find the Pillars of Strength (PS) heads that constitute the core of the GHA and remove other heads. 
We first describe the V2S roughly. In GHA, the heads within each group exhibit similar patterns upon the convergence of the Group-Constrained Training (GCT). 
Then, we only keep the heads with the most explicit group patterns (the PS heads), and switch off the other ones within the same group via V2S. The main idea of V2S is to vote on all heads of the GHA, and only retain one head for each group -- the head receiving the most votes. Specifically, it takes an entire epoch to collect the layer-wise votes \(\*m_l^b \in \{0, 1\}^k\) from the whole training set (each data batch $b$ creates one layer-wise vote $\*m_l^b$ per attention layer), where $k$ denotes the head number; \(0\) indicates that the corresponding head should be switched off and \(1\) indicates that a head is retained. 

We assume that there are \(B\) mini-batches in the training set. Then, each attention layer receives \(B\) layer-wise votes within which each head-wise vote is denoted by either \(0\) or \(1\). For each  group, the head receiving the most `\(1\)'s are assigned a `\(1\)' in the final head mask \(\*m_l \in \{0, 1\}^k\) for attention layer \(l\), indicating that this head will be retained. Following~\citet{DBLP:conf/nips/MichelLN19} and~\citet{DBLP:conf/acl/VoitaTMST19}, we mask out the output of heads as the equivalent operation of head removal\footnote{We perform real head removal when test inference speed.}.
\begin{algorithm}[t]
\caption{The Voting-to-Stay (V2S) algorithm}
\label{Voting to Stay}
\begin{algorithmic}[1]
\State \textbf{Procedure} Voting-to-Stay($\*f, \hat{\*V}, \*A, \*O, \*Z$)
	\If {satisfy \(\*\rho\), and \(\*m\) is none}
    	\State Start voting epoch; Freeze \(\*f\).
        \State \(\*\Gamma_l\) \(\gets [ \,\ ] \,\) \textcolor{teal}{\Comment{Creat \(\*\Gamma_l\) to store votes}}
        \For {batch \(b\)  in \(B\) training batches}
            \For {layer \(l\) in \(L\) layers}
        	\For {\(\*E_l\) in \(\{\hat{\*V}_{l}, \*A_{l}, \*O_{l}\}\) }
            	\State Based on $\eta_l = \{\eta_{1, l}, ..., \eta_{1, k}\}$,
                    \State create \(\*m_{l, v}^b, \*m_{l, a}^b, \*m_{l, o}^b\).
            \EndFor
            \State Add \(\*m_{l, v}^b\), \(\*m_{l, a}^b\), \(\*m_{l, o}^b\) to \(\*\Gamma_l\).
            \EndFor
        \EndFor
        \For {\(l\) in \(n\)} \textcolor{teal}{\Comment{Vote at each attn layer}}
        	\State \(\*m_l \gets VOTE(\*\Gamma_l)\) 
        \EndFor
        \State \(\*m\gets [\*m_1, ..., \*m_n]\) \textcolor{teal}{\Comment{Stack layer votes}}
        \State Unfreeze \(\*f\); end voting epoch.
    \EndIf
  \State \(\*f = \*f \odot \*m\)  \textcolor{teal}{\Comment{Mask GHT attn outputs with \(\*m\)}}

\end{algorithmic}
\end{algorithm}
The V2S procedure is outlined in Algorithm~\ref{Voting to Stay}. We detail some of its definitions below. \textbf{(1)} \(\*\rho\) indicates the full convergence of GHT, i.e., the hidden units found by \(\Omega\) have a center shift less than a threshold. \textbf{(2)} In Step 7-9, given feature maps \(\hat{\*V}_{l}\), \(\*A_{l}\), and \(\*O_{l}\) of the \(l\)-th attention layer, the vote vectors \(\*m_{l, v}^b\), \(\*m_{l, a}^b\), and \(\*m_{l, o}^b \in \{0, 1\}^k\) are determined by the group pattern scores \(\*\eta_{i,l}\) of each head, indicating the explicitness of group patterns. 

We set the corresponding digit in the vote vectors as 1 for the head achieving the highest $\eta_{i,l}$ in its group, indicating the most representative head of the group. Here \(\*\eta_{i,l} = p_z(\*z_{i, l} | \*e_{i,l})\) if \(\*z\) is categorical; otherwise \(\*\eta_{i,l} = -\varphi(\*e_{i, l}; \*z_{i, l})\). \(\*e_{i,l}\) denotes the $i$-th head feature map (either one of the \(\*v_{i, l}\), \(\*a_{i, l}\), or \(\*o_{i, l}\)). 
\textbf{(3)} \(VOTE\) means counting the `\(1\)'s for each head based on the \textit{0-1} votes in \(\*\Gamma_l\) and only keeping the heads with the highest counts\footnote{Besides voting, there is an alternative way to create the mask. Instead of using \textit{0-1} number as a discrete voting unit, the group pattern scores can be added up to rank the head pattern explicitness. We find that the two ways perform similarly.}. After V2S, a finetuning is applied to adapt the pruned network.

GHT-PS compresses considerable parameters. In the case of two head groups, GHT-PS reduces 75\% parameters for an attention layer and 32.1\% for the entire model\footnote{The encoder-decoder arch in~\citet{DBLP:conf/nips/VaswaniSPUJGKP17}.}. We will show that V2S removing non-PS heads does not sacrifice model performance. Instead, it brings accuracy gains in some cases and improves inference speed.


\section{Experimental Setup}
\label{Experimental Setup}
In this section, we detail the key architectural configurations. Further training, model, dataset \& evaluation setups are detailed in~\ref{Trainig Settings},~\ref{GHT model setting}, \&~\ref{Datasets and Evaluation}.
We follow the transformer of~\citet{DBLP:conf/nips/VaswaniSPUJGKP17} as a backbone architecture for all datasets and tasks in our experiments. 
Following~\citet{DBLP:conf/iclr/WuFBDA19,DBLP:conf/iclr/WuLLLH20}, for Machine Translation and Abstractive Summarization, we adopt the same 8-head encoder-decoder architecture with 6 layers for both encoder and decoder, where the model dimension \(d_{model}\) $=$ \(512\) and feed-forward dimension \(d_{f}\) $=$  \(2048\). For LM, we adopt the 16-head decoder-only architecture with 16 layers, where the model dimension \(d_{model}\) $=$ \(1024\) and feed-forward dimension \(d_{f}\) $=$  \(4096\).  The layer normalization is applied before the residual connection of each layer. The parameters of decoder input and output projections are shared. Our models are based on fairseq~\citep{ott-etal-2019-fairseq}
implementations. 

We perform the GCT as a metric learning task because it does not introduce additional projection layers when the shapes of similarity inputs are identical (Eq.\ref{eq: grouping}(B)), which makes GHT weight-lighter. In addition, it performs better in our experiments compared to the classification-based grouping. 

\begin{table*}[t]
\small
\begin{center}
\begin{tabular}{lllllll@{\hskip 0.15in}ll@{\hskip 0.3in}ll}
\toprule
\multirow{2}{*}{Model} & \multirow{2}{*}{Param \(\downarrow\)} & \multirow{2}{*}{\begin{tabular}{c}Inference\\Speed \(\uparrow\)\end{tabular}} & \multirow{2}{*}{FLOPs \(\downarrow\)} & \multicolumn{2}{l}{BLEU \(\uparrow\)} &\multicolumn{3}{l}{\textit{\!\!IWSLT}}             & \multicolumn{2}{c}{\textit{\!\!WMT}} \\ \cmidrule{5-9} \cmidrule{10-11}
                        &                 &  &            & de-en & it-en & en-de & en-it & en-fr & en-de      & en-fr      \\ \midrule
Vanilla Transformer & 44M    & 1012.1 sent/s &  1996M                  & 34.4  & 32.3  & 28.0  & 30.8  & 40.1  & 27.3       & 38.1       \\\cmidrule{2-11}
GHT (ours)                     & 44M      & 1016.4 sent/s &      1996M            & \textbf{35.4}  & \textbf{32.8}  & \textbf{29.1}  & \textbf{31.6}  & \textbf{41.5}  & \textbf{28.6}       & \textbf{40.7}       \\ \midrule
Transformer-Lite1 & 30M         & 1175.4 sent/s &    1549M           & 33.8  & 31.9  & 27.9  & 29.3  & 39.9  & 26.9       & 37.7       \\
Transformer-Lite2 & 30M         & 1108.7 sent/s &     1465M          & 34.0  & 32.2  & 28.2  & 29.5  & 40.0  & 26.7       & 37.8       \\\cmidrule{2-11}
GHT-PS (ours)                  & 30M       & 1122.1  sent/s &     1558M            & \textbf{35.2}  & \textbf{32.7}  & \textbf{28.9}  & \textbf{31.6}  & \textbf{41.4}  & \textbf{28.2}       & \textbf{40.5}       \\ \bottomrule
\end{tabular}
\caption{Benchmark with vanilla transformer (backbone) on IWSLT and WMT Machine Translation datasets, measured by BLEU. All improvements are statistically significant with $p < 0.05$ under t-test.
}
\label{BLEU-transformer-GBT}
\end{center}
\end{table*}
\begin{table*}[t]
\setlength{\tabcolsep}{4.3pt}
\small
\begin{center}
\begin{tabular}{lllllllll@{\hskip 0.3in}cc}
\toprule
\multirow{2}{*}{Model} & \multirow{2}{*}{Param \(\downarrow\)} & \multirow{2}{*}{\begin{tabular}{c}Inference\\Speed \(\uparrow\)\end{tabular}} & \multirow{2}{*}{FLOPs \(\downarrow\)} & \multicolumn{2}{l}{BLEU \(\uparrow\)} &\multicolumn{3}{l}{\textit{\!\!IWSLT}}             & \multicolumn{2}{c}{\textit{\!\!WMT}} \\ \cmidrule{5-9} \cmidrule{10-11}
                        &              &  &              & de-en & it-en & en-de & en-it & en-fr & en-de      & en-fr      \\ \midrule
\citet{DBLP:journals/corr/abs-2006-16362}                    & 44M        & 416.6 sent/s &     2054M       & 34.4  & 31.8  & 28.2  & 31.0  & 40.7  & 27.6       &   38.5    \\
\citet{DBLP:conf/emnlp/LiTYLZ18}                  & 44M        & 1011.2 sent/s &     1996M       & 34.7  & 31.8  & 28.5  & 30.7  & 40.7  & 27.3       &     39.3   \\\cmidrule{2-11}
GHT (ours)                     & 44M         & 1016.4 sent/s &         1996M      & \textbf{35.4}*  & \textbf{32.8}*  & \textbf{29.1}*  & \textbf{31.6}*  & \textbf{41.5}*  & \textbf{28.6}*       & \textbf{40.7}*       \\ \midrule
\citet{DBLP:conf/acl/VoitaTMST19}                   & 30M     & 1099.1 sent/s &        1558M       & 32.2  & 30.8  & 26.5  & 30.3  & 39.8  & 22.0       & 34.0       \\
\citet{DBLP:journals/tacl/LiCS21} 
& 30M      & 1116.9 sent/s &       1558M       & 33.2  & 31.3  & 27.5  & 30.0  & 39.7  & 20.5       & 33.6       \\
Dynamic   conv 
& 30M           & 1050.2 sent/s &    1615M     & 34.8  & \textbf{32.7}  & 28.7  & 31.1  & 40.6  & 24.0       & 36.5       \\
Lite   Transformer 
& 30M           & 1096.6 sent/s &    1809M     & 33.3  & 31.4  & 27.5  & 29.8  & 39.4  & 24.9       & 37.4       \\ \cmidrule{2-11}
GHT-PS (ours)                  & 30M           & 1122.1   sent/s &     1558M      & \textbf{35.2}*  & \textbf{32.7}  & \textbf{28.9}*  & \textbf{31.6}*  & \textbf{41.4}*  & \textbf{28.2}*       & \textbf{40.5}*      \\ \bottomrule  
\end{tabular}
\caption{Benchmark with state-of-the-art MHA redundancy/parameter optimization baselines on IWSLT and WMT Machine Translation datasets at the same parameter level, measured by BLEU. * denotes the improvement is statistical significant with $p < 0.05$ under t-test.
}
\label{BLEU-SOTA-GHT-PS}
\end{center}
\end{table*}

\section{Results and Analysis}
\label{Results}
\subsection{Machine Translation}

\paragraph{Ours vs. vanilla transformer.} We first report results by comparing GHT and GHT-PS with the vanilla transformer~\citep{DBLP:conf/nips/VaswaniSPUJGKP17} which is the backbone of our model. As shown in Table~\ref{BLEU-transformer-GBT}, the models are compared at different parameter levels\footnote{The parameters analyzed in this paper exclude the embedding layer since they vary a lot between different datasets when the vocabulary sizes are different.}. GHT does not have weight reduction, keeping the same parameter size as the vanilla transformer (44M, the same setting as transformer base 
~\citep{DBLP:conf/nips/VaswaniSPUJGKP17}). In contrast, GHT-PS is compressed to 30M parameters via V2S.
For a fair comparison, we first compare GHT-PS with two lite architectures, Transformer-Lite1 and Transformer-Lite2, whose parameter numbers are 30M as well. Keeping other settings unchanged, the encoder and decoder of Transformer-Lite1 are reduced to 4 layers, respectively. Transformer-Lite2 reduces the model dimension \(d_{model}\) to 424, and \(d_{f}\) to 1696. 

GHT and GHT-PS consistently and significantly outperform their backbone models at the same parameter level across all datasets. On average, the GHT surpasses 44M vanilla transformer by 3.8\% in BLEU ~\cite{DBLP:conf/acl/PapineniRWZ02}; GHT-PS surpasses Lite1 and Lite2 by 4.9\% and 4.4\%, respectively. Although GHT-PS reduces 32.1\% parameters, it significantly outperforms both 44M and 30M vanilla transformers, which is comparable to GHT on all datasets. It shows that V2S reduces the parameter size of the original transformer without sacrificing accuracy on MT. Efficiency is analyzed later.


\paragraph{Ours vs. efficient attention models.} We compare GHT with two state-of-the-art (SOTA) MHA redundancy optimization baselines. \citet{DBLP:journals/corr/abs-2006-16362} and~\citet{DBLP:conf/emnlp/LiTYLZ18} are respectively homogenization- and diversification-based methods. In addition, we compare GHT-PS with four SOTA baselines that made major contributions to attention parameter compression and redundancy optimization\footnote{Works optimizing parameters of transformer modules rather than the MHA are not compared. In addition, we do not compare to~\citet{DBLP:conf/nips/MichelLN19} (post-pruning), because their method performs extremely bad when the parameter level is low, e.g., 30M~\citep{DBLP:journals/tacl/LiCS21}.}. ~\citet{DBLP:conf/acl/VoitaTMST19} and~\citet{DBLP:journals/tacl/LiCS21} are significance-based pruning methods. Dynamic Conv~\citep{DBLP:conf/iclr/WuFBDA19} and Lite Transformer~\citep{DBLP:conf/iclr/WuLLLH20} modify the MHA arch to reduce parameters. 

Table~\ref{BLEU-SOTA-GHT-PS} shows that GHT outperforms all its baselines on all datasets, exceeding the strongest baseline by 2.9\% in averaged BLEU scores. GHT-PS outperforms all its baselines on 6 out of 7 datasets, exceeding the strongest baseline by 4.4\% on average. Model compression of the baselines may sacrifice performance (especially on large datasets, e.g., WMT en-de and en-fr), while GHT-PS is almost not affected by the parameter reduction, even surpassing GHT's baselines with 44M parameters. 




\begin{table}[h]
\setlength{\tabcolsep}{5.5pt}
\small
\begin{center}
\begin{tabular}{@{}l@{\hskip 0.1in}ccccc@{}}
\toprule
\multirow{2}{*}{Model}                        & \multicolumn{5}{c}{BLEU \(\uparrow\)}\\\cmidrule{2-6} & de-en         & it-en         & en-de         & en-it         & en-fr         \\ \midrule
GHT                          & \textbf{35.4} & \textbf{32.8} & \textbf{29.1} & \textbf{31.6} & \textbf{41.5} \\ 
- w/o Diversifying                 & 34.7          & 31.8          & 28.5          & 30.7          & 40.7          \\
- w/o Homologizing                 & 34.3          & 32.0          & 28.2          & 30.9          & 40.2          \\ \midrule
GHT-PS                       & \textbf{35.2} & \textbf{32.7} & \textbf{28.9} & \textbf{31.6} & \textbf{41.4} \\
- w/o GCT        & 33.8          & 31.9          & 28.1          & 30.5          & 39.8          \\
- w/o GC & 34.0          & 32.0          & 28.4          & 31.0          & 40.2          \\
- w/o HUDS              & 33.7          & 32.0          & 28.1          & 30.9          & 40.3          \\
- w/o PS stay  & 33.6          & 31.7          & 27.9          & 30.7          & 40.2          \\
- w/ stage 2 GC            & 33.2          & 31.8          & 28.1          & 30.8          & 40.3          \\
- w/ stage 1\& 2 GC              & 33.4          & 31.9          & 27.7          & 30.6          & 40.2          \\ \bottomrule
\end{tabular}
\caption{Ablation study on IWSLT'14. The results are statistically significant with $p < 0.05$ under t-test.}
\label{Ablation}
\end{center}
\end{table}

\paragraph{Ablation Study.} We evaluate the impacts of the features we choose for GHT and GHT-PS (Table~\ref{Ablation}). We first ablate the diversification/homogenization term of GCT (see Eq.\ref{eq: grouping}), which lowers the BLEU scores. 
Next, we show the importance of GCT for V2S. \textbf{w/o GCT} denotes that we directly perform V2S at the very beginning without GCT. \textbf{w/o GC} denotes that V2S is employed after normal training without Group Constrain (GC). Both ablation models yield lower BLEU, because they do not homogenize unnecessary heads and prepare them for pruning.
Next, we validate the power of Pillars of Strength. \textbf{w/o HUDS} denotes we replace HUDS with randomly switching off heads after GCT; \textbf{w/o PS stay} denotes we keep random group members instead of the Pillars of Strength after GCT. We observe lower BLEU in \textbf{w/o HUDS} and \textbf{w/o PS stay}. Finally, we find that GC only needs to be added before V2S. We denote the training stages before and after V2S as stages 1 and 2. We compare the proposed Stage 1-based GHT-PS with models that perform GCT at Stage 2 (\textbf{w/ stage 2 GC}) and at both stages (\textbf{w/ stage 1\& 2 GC}). BLEU scores of both ablation models decrease.

\begin{figure}[t]
\centering
\includegraphics[width=0.41\textwidth]{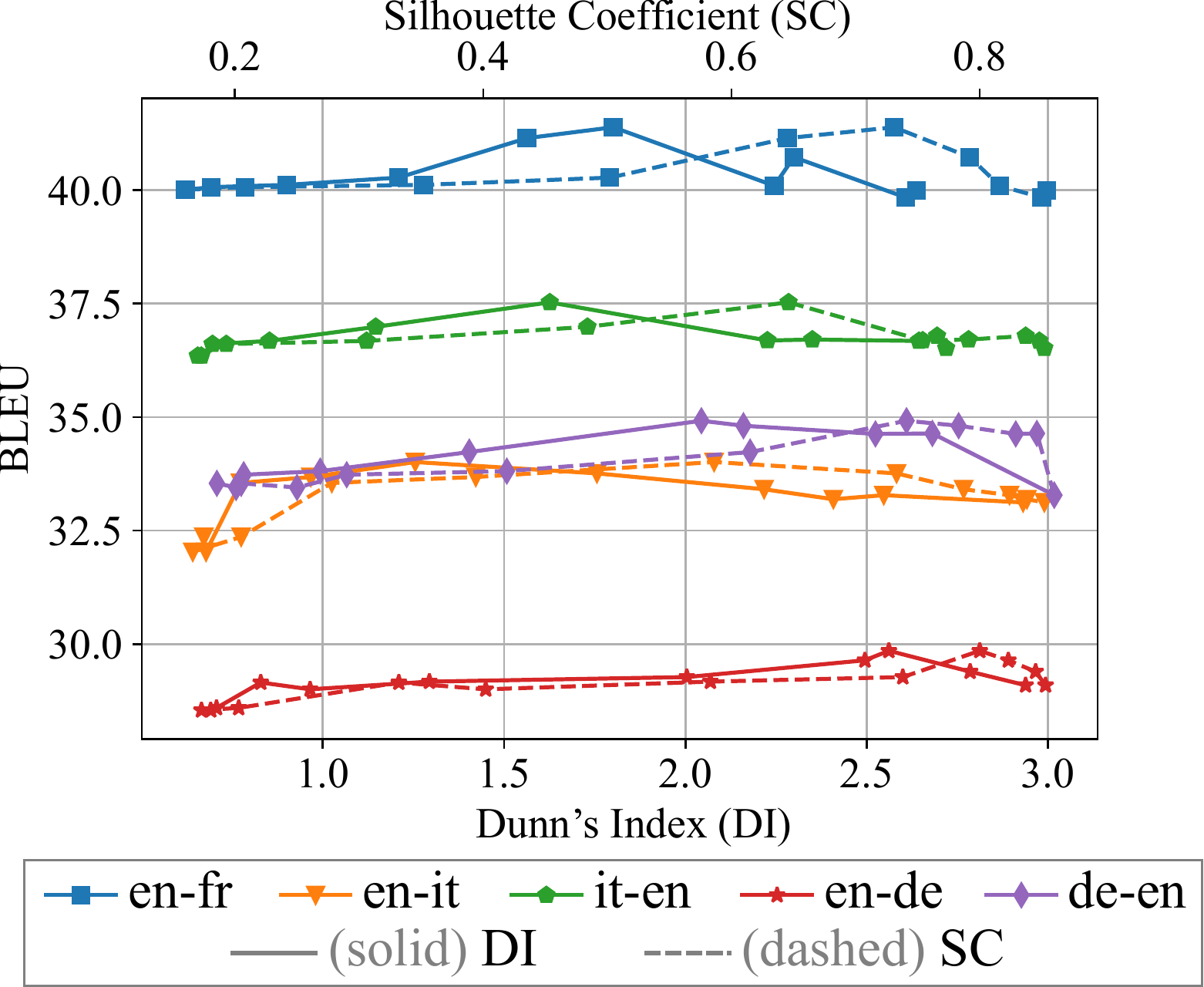}
\caption{The final BLEU scores achieved by GHT (on IWSLT'14 dev set) first rise and then drop, as the final group patterns become more compact (indicated by the increasing SC and DI scores).}
\label{BLEU-DI-SC}
\end{figure}

\paragraph{Effect of group compactness.} We hypothesize that more compact group patterns bring performance gains to the GHT. Figure~\ref{BLEU-DI-SC} shows the correlation between the compactness of the final group patterns and the final BLEU scores GHT achieved on 5 IWSLT’14 development sets when the GHT is fully converged in GCT. One data point corresponds to an independent run. We choose Silhouette Coefficient (SC)~\citep{rousseeuw1987silhouettes} and Dunn's Index (DI)~\citep{bezdek1995cluster} as the measurements of group pattern compactness, both of which increase as the intra-group samples become more similar and the inter-group ones become more separated. The SC and DI are computed with the FMs of GHA (\S~\ref{sec:GHA}) and controlled by tuning the $\alpha$ and $\beta$ (Eq.\ref{eq: grouping}). \\

Figure~\ref{BLEU-DI-SC} shows that, within the normal range, the BLEU scores rise with higher SC/DI scores, which is in line with our assumption. The BLEUs start to drop after the peak as the SC/DI scores increase, because the very heavy group constraint prohibits the model from learning useful task-specific knowledge.

\begin{figure}[t]
\centering
\includegraphics[width=0.41\textwidth]{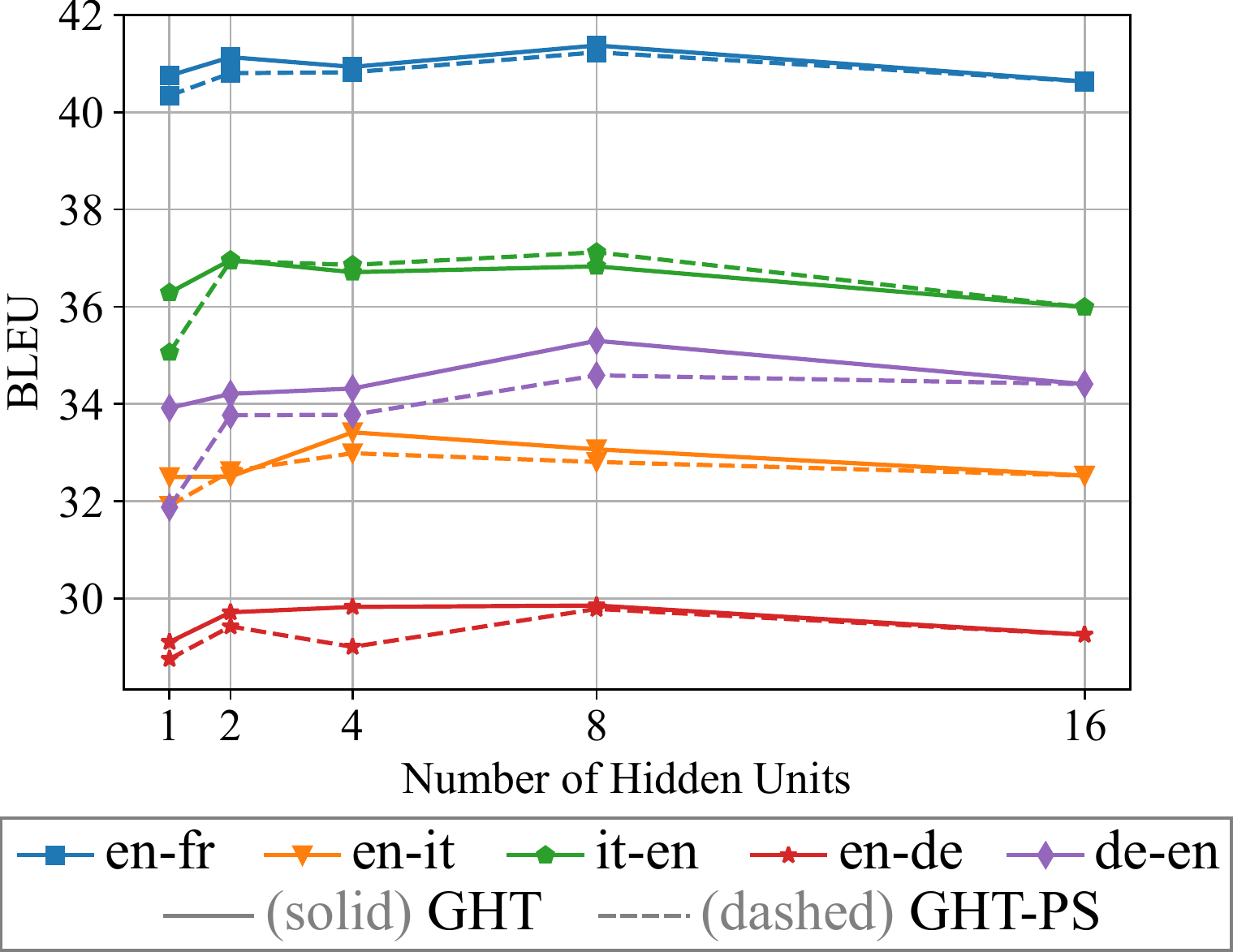}
\caption{The BLEUs of GHT and GHT-PS by different numbers of hidden units (groups) on IWSLT'14 dev set.}
\label{Group number analysis}
\end{figure}

\begin{table*}[t]
\setlength{\tabcolsep}{5pt}
\renewcommand{\arraystretch}{0.95}
\small
\begin{center}
\begin{tabular}{lcccccc}
\toprule
Model                  & Param \(\downarrow\) & Inference   Speed \(\uparrow\) & FLOPs \(\downarrow\) & Rouge-1 \(\uparrow\) & Rouge-2 \(\uparrow\) & Rouge-L \(\uparrow\)  \\ \midrule
LSTM~\citep{DBLP:conf/iclr/PaulusXS18}                  & -   & -                 & -  & 38.30   & 14.81   & 35.49        \\
CNN~\citep{DBLP:conf/aclnmt/FanGA18}                   & -   & -                 & -  & 39.06   & 15.38   & 35.77        \\
Light Conv~\citep{DBLP:conf/iclr/WuFBDA19}              & 86M & -                 & -  & 39.52   & 15.97   & 36.51        \\
Dynamic Conv~\citep{DBLP:conf/iclr/WuFBDA19}            & 87M & -                 & -  & 39.84   & 16.25   & 36.73        \\ \midrule
Vanilla Transformer~\citep{DBLP:conf/acl/VoitaTMST19} & 44M & 208.77   sent/s  & 1996M  & 38.45   & 17.97   & 36.03    \\\midrule
GHT (ours)          & 44M  & 208.77   sent/s  & 1996M & 40.00    & 21.10    & 37.51     \\
GHT-PS (ours)   & \textbf{30M} & \textbf{257.62 sent/s}  & \textbf{1558M}  & \textbf{40.01 }   & \textbf{21.31}    & \textbf{37.62}     \\ \bottomrule
\end{tabular}
\caption{Abstractive Summarization results on CNN-DailyMail in terms of F1-Rouge and efficiency (parameter, inference speed, and FLOPs). All improvements are statistically significant with $p < 0.05$ under t-test.}
\label{Abstractive Summarization}
\end{center}
\end{table*}

\paragraph{Effect of group number.} Figure~\ref{Group number analysis} shows the performance trends of 16-head GHT and GHT-PS by different numbers of group hidden units. For GHT, different datasets have different optimal hidden unit quantities, while a similar trend is observed. The optimal group number is between 2 and 8, which is in line with the claim that our group strategy is superior to sole homogenization (1 group) or diversification (16 groups) strategies. For GHT-PS, when the group number is larger than 1, it shows comparable performance to GHT on most datasets. This also verifies that non-PS heads can be switched off without sacrificing performance.

\begin{figure}[h]
\centering
\includegraphics[width=0.41\textwidth]{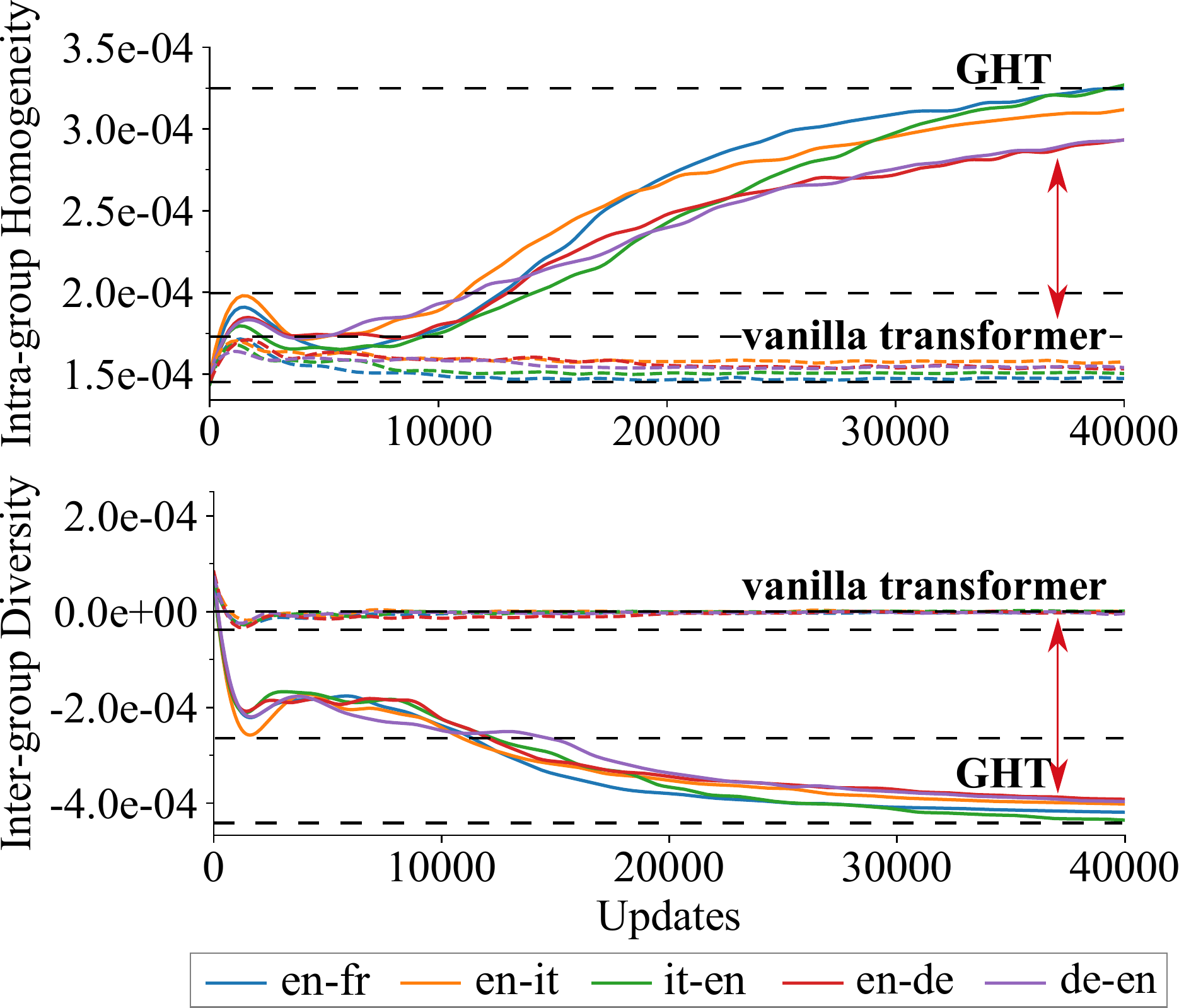}\vspace{-0.5em}
\caption{Intra-group homogeneity (upper) and inter-group diversity (lower) of GHT and vanilla transformer by training steps.}
\label{fig: Intra and inter degree}
\end{figure}

\paragraph{Group pattern trends.} Figure~\ref{fig: Intra and inter degree} shows the trends of intra-group homogeneity (given by the \textit{1st} term of Eq.\ref{eq: grouping}(B)) and inter-group diversity (given by the \textit{2nd} term of Eq.\ref{eq: grouping}(B)) of GHT and vanilla transformer in the training process on five IWSLT datasets.
By training, GHT yields higher intra-group homogeneity and inter-group diversity absolute values, leading to more compact groups, while the vanilla transformer shows flattened trends. It shows that GCT can effectively homogenize intra-group heads and diversify inter-group heads.

\begin{table}[h]
\setlength{\tabcolsep}{1.5pt}
\renewcommand{\arraystretch}{0.95}
\small
\begin{center}
\begin{tabular}{@{}l@{}cccc@{}}
\toprule
Model             & BLEU\(\uparrow\) & Param\(\downarrow\) & Infer Speed\(\uparrow\) & FLOPs\(\downarrow\) \\ \midrule
Transformer base & 25.8 & 44M       & 1016.4   sent/s & 1996M \\
Transformer big    & 26.4 & 176M      & 707.1   sent/s  & 6635M \\
Lite   Conv        & 26.6 & 166M      & 722.1   sent/s  & 6184M \\
Dynamic Conv     & 26.9 & 176M      & 710.0   sent/s  & 6315M \\ \midrule
GHT-PS-LITE (ours)        & 26.6 & \textbf{16M}       & \textbf{1170.2}   sent/s & \textbf{1181M} \\ 
\bottomrule
\end{tabular}
\caption{Efficiency comparison by parameter, inference speed (averaged on five runs), and FLOPs. All results are generated with beam size 5, batch size 256, max decoding length 10 on a NVIDIA Quadro RTX A6000.}
\label{Efficiency-Analysis}
\end{center}
\end{table}

\paragraph{Efficiency analysis.} In Tables~\ref{BLEU-transformer-GBT} and~\ref{BLEU-SOTA-GHT-PS}, the efficiency metrics 
are controlled to be identical. Our models with higher inference speed and lower FLOPs show efficiency by culling redundant parameters. 
We also compare the efficiency metrics by controlling BLEU scores. In Table~\ref{Efficiency-Analysis}, we select models from the works in Table~\ref{BLEU-transformer-GBT} and~\ref{BLEU-SOTA-GHT-PS} that are reported to achieve close BLEU scores on newstest2013 as the baselines.
The GHT-PS-LITE is a light version of GHT-PS that has a \(d_{f}\) of 1024. 
Given BLEU ranges from 25.8 to 26.9, GHT-PS-LITE is much more efficient than the baselines. Noticeably, GHT-PS-LITE achieves 90.36\% fewer parameters, 62.05\% faster inference speed, and 80.90\% fewer FLOPs against Lite Conv which yields the same BLEU as it.

\subsection{Abstractive Summarization}

We evaluate the ability of our model to process longer inputs via Abstractive Summarization on the CNN-DailyMail dataset. We take vanilla transformer as the backbone. 
Table~\ref{Abstractive Summarization} shows that both GHT and GHT-PS achieve higher F1-Rouge scores ~\cite{lin2004rouge} on this task. GHT-PS achieves 4.1\% higher Rouge-1, 18.6\% higher Rouge-2, and 4.4\% higher Rouge-L against vanilla transformer. It also achieves 0.4\% higher Rouge-1, 31.1\% higher Rouge-2 and 2.4\% higher Rouge-L against the best-performing baseline (Dynamic Conv). Meanwhile, GHT-PS only takes 68.18\% parameters of the vanilla transformer and exhibits higher inference speed and fewer FLOPs.\\

\begin{table}[h]
\setlength{\tabcolsep}{2.5pt}
\renewcommand{\arraystretch}{0.95}
\small
\begin{center}
\begin{tabular}{@{}l@{\hskip 0.02in}ccccc@{}}
\toprule
Model                           & Param\(\downarrow\) & Infer   Spd\(\uparrow\) & FLOPs\(\downarrow\) & Valid\(\downarrow\) & Test\(\downarrow\)   \\ \midrule
S4                              & 249M & -                 & - & 19.69 & 20.95      \\
BERT-L-CAS                 & 395M & -                 & - & 19.67 & 20.42      \\
GPT-2   Large                  & 762M & -                 & - & -     & 22.05      \\ \midrule
VT w/ AI              & 201M & 9.9   tok/s   & 6106M & 19.03 & 19.14  \\ \midrule
GHT (ours)                   & 201M & 9.9   tok/s   & 6106M & \textbf{18.57} & 18.60  \\
GHT-PS (ours)            & \textbf{167M} & \textbf{19.0   tok/s}  & \textbf{4573M} & 18.58 & \textbf{18.59}  \\ \bottomrule
\end{tabular}
\caption{Language modeling results on WIKITEXT-103 by perplexity and efficiency (parameter, inference speed, and FLOPs). VT w/ AI denotes vanilla transformer with adaptive input. All improvements against the baselines are statistically significant with $p < 0.05$ under t-test.}
\label{Language modeling}
\end{center}
\end{table}

\subsection{Language Modeling}
We evaluate LM performance on WIKITTEXT-103 dataset. The backbone is a decoder-only transformer with 16 layers and adaptive inputs~\citep{DBLP:conf/iclr/BaevskiA19}. We compare with the backbone model, as well as comparable SOTA LM models, including S4~\citep{DBLP:conf/iclr/GuGR22}, BERT-Large-CAS~\citep{DBLP:journals/corr/abs-1904-09408}, and GPT-2 Large~\citep{noauthororeditor}.

Table~\ref{Language modeling} shows that both GHT and GHT-PS achieve lower perplexity ~\cite{DBLP:journals/corr/Vajapeyam14} than the baselines on both validation and test sets (2.9\% and 9.0\% less perplexity against the backbone and the best performing LM baseline, respectively). Meanwhile, GHT-PS achieves 16.92\% parameter reduction, $2$ times faster inference speed, and 75\% FLOPs compared with the backbone.

\section{Conclusion}
In this paper, we assume that only focusing on the most representative and distinctive features with minimum resources may mitigate the redundancy and over-parameterization issues of MHA. Accordingly, we propose a divide-and-conquer strategy, including GCT and V2S to mitigate the issues.
The improvements on three tasks and the extensive analysis verify our hypothesis and the effectiveness of our redundancy optimization methods. Our study may inspire future MHA design and training to achieve higher accuracy and efficiency.

\section*{Limitations}
In this work, we evaluate the proposed models for NLP tasks only. However, tasks in other fields such as Computer Vision may present a very different input inductive bias, thus affecting the performance. Moreover, our models are trained from scratch, hence it is unknown whether the same divide-and-conquer strategy works for pre-trained models. We will study these limitations in the future to give a more extensive exploration.

\section*{Ethics Statement}
This article follows the ACL Code of Ethics. The annotations are based on public datasets that do not contain private data. The algorithm we developed is an architectural optimization technique for improving model performance. To our best knowledge, there are no foreseeable potential risks to using this technique.

\section*{Acknowledgments}
This research is supported by the Agency for Science, Technology and Research (A*STAR) under its AME Programmatic Funding Scheme (Project \#A18A2b0046).




\bibliography{custom}
\bibliographystyle{acl_natbib}

\clearpage



\appendix
\section{Appendix}

\subsection{Trainig Settings}
\label{Trainig Settings}
\subsubsection{Machine Translation}
We use Adam to optimize the MT models and set the \(\beta_1 = 0.9, \beta_2 = 0.98\). We use the Inverse Square Root Schedule~\citep{DBLP:conf/nips/VaswaniSPUJGKP17} where it first warms up for 4K steps until the learning rate reaches \(5 \times 10^{-4}\), and then it exponentially decays the learning rate. We apply early stop as a termination condition. We apply a 0.3 dropout rate for all Machine Translation models. A weight decay of \(10^{-4}\) is used for all IWSLT 2014 models, whereas for WMT models we use a weight decay of 0. We apply a 0.1 label smoothing~\citep{DBLP:conf/cvpr/SzegedyVISW16, DBLP:conf/iclr/PereyraTCKH17} for the uniform prior distribution over the vocabulary. 

\subsubsection{Language Modeling}
Following~\citet{DBLP:conf/iclr/BaevskiA19}, we use Nesterov's accelerated gradient method~\citep{DBLP:conf/icml/SutskeverMDH13} with a momentum of 0.99. We clip the gradient norm if it exceeds 0.1~\citep{DBLP:conf/icml/PascanuMB13}. The learning rate is linearly warmed up from \(10^{-7}\) to 1 for 16K steps and then annealed using a cosine learning rate schedule~\citep{DBLP:conf/iclr/LoshchilovH17} with multiple cycles. Each cycle doubles the number of updates than the previous cycle and we shrink the maximum and minimum learning rates by 0.75 compared to the previous cycle. The initial minimum learning rate is \(10^{-4}\) and the maximum is 1. We apply 0.2 adaptive softmax dropout rate, 0.1 attention dropout rate, and 0.1 activation dropout rate. 

\subsubsection{Abstractive Summarization} 
We use the same training setup with IWSLT 2014 models. We apply 0.1 clip norm and 0.2 attention dropout. The model is warmed up for 10K updates.

\subsection{Further Model Settings}
\label{GHT model setting}
Different \(\alpha\), \(\beta\), and head feature maps (\(\hat{\*V}\), \(\*A\), and \(\*O\)) are preferred for different datasets to achieve optimal performance.
The Machine Translation configurations are detailed in Table~\ref{tab:MTconfig}; Language Modeling and Abstractive Summarization configurations are detailed in Table~\ref{tab:LMASconfig}.

Note that \(\varphi(\*v_{i, l}, \*a_{i, l}, \*o_{i, l}; \*z_{i, l})\) \(=\) \(\tau_1 \varphi(\*v_{i, l}\); \(\*z_{i, l})\) \(+\) \(\tau_2 \varphi(\*a_{i, l}\); \(\*z_{i, l})\) \(+\) \(\tau_3 \varphi(\*o_{i, l}\); \(\*z_{i, l})\), we set one of the $\{\tau_1, \tau_2, \tau_3\}$ to be 1, the other to be 0.

\subsection{Datasets and Evaluation\footnote{For all three tasks, we follow the data pipeline of fairseq: \url{https://github.com/facebookresearch/fairseq/blob/main/examples}}}
\label{Datasets and Evaluation}

\subsubsection{Efficiency Metrics settings}
\label{Efficiency Metrics settings}
\paragraph{Inference speed.} All inference speed results are generated with beam size 5, batch size 256, maximum decoding length 10 on a single NVIDIA Quadro RTX A6000.

\paragraph{FLOPs.} We use the fvcore\footnote{\url{https://github.com/facebookresearch/fvcore}} to calculate the FLOPs, with a fixed input length of 30. 

\subsubsection{Machine Translation} 
To fully evaluate the effectiveness of our methods, we evaluate seven MT datasets of IWSLT'14 and WMT 2014 benchmarks. We closely follow the setup of~\citet{DBLP:conf/nips/VaswaniSPUJGKP17} for data preparation. WMT 2014 English-German dataset consists of about 4.5M sentence pairs. It is encoded with byte-pair encoding~\citep{DBLP:journals/corr/BritzGLL17}, having a shared source-target vocabulary of about 40K tokens. Following the standard setting~\citep{DBLP:conf/nips/VaswaniSPUJGKP17}, we validate on newstest2013 and test on newstest2014 for experiments on this dataset. The WMT 2014 English-French dataset consists of 36M sentence pairs and is encoded with a joint source-target BPE of about 43K vocabularies. Following the standard split, we validate on a concatenation of newstest2012 and newstest2013 and test on newstest2014. For IWSLT’14 German to English, IWSLT’14 English to German, IWSLT’14 English to French, IWSLT’14 English to Italian and IWSLT’14 Italian to English, we encode the sentence pairs with joint source-target BPE. Following~\citet{DBLP:conf/naacl/EdunovOAGR18}, the validation set is randomly splited from the training set with a ratio of 1:23. The testset consists TED.tst2010, TED.tst2011, TED.tst2012 and TED.dev2010, TEDX.dev2012 for IWSLT’14 German to English, IWSLT’14 English to German, and IWSLT’14 English to French; the TEDX.dev2012 is replaced by TEDX.dev2014 for IWSLT’14 English to Italian and IWSLT’14 Italian to English.

For all Machine Translation datasets, we use detokenized BLEU.
WMT 2014 English-German and WMT 2014 English-French are evaluated with a beam size 4 and length penalty 0.6; IWSLT’14 datasets are evaluated with a beam size 5 and without length penalty.

\begin{table*}[t]
\centering
\begin{tabular}{|c|ccccc|cc|}
\hline
\multirow{2}{*}{Model} & \multicolumn{5}{c|}{IWSLT (\(\alpha/\beta/FM\))}                                                                                                    & \multicolumn{2}{c|}{WMT (\(\alpha/\beta/FM\))}   \\ \cline{2-8} 
                       & \multicolumn{1}{c|}{de-en}     & \multicolumn{1}{c|}{it-en}     & \multicolumn{1}{c|}{en-de}     & \multicolumn{1}{c|}{en-it}     & en-fr     & \multicolumn{1}{c|}{en-de}     & en-fr     \\ \hline
GHT                    & \multicolumn{1}{c|}{0.7/0.5/\(\hat{\*V}\)} & \multicolumn{1}{c|}{0.3/0.5/\(\hat{\*V}\)} & \multicolumn{1}{c|}{0.3/0.1/\(\*A\)} & \multicolumn{1}{c|}{0.3/0.3/\(\hat{\*V}\)} & 0.7/0.7/\(\hat{\*V}\) & \multicolumn{1}{c|}{0.5/0.5/\(\hat{\*V}\)} & 0.3/0.3/\(\hat{\*V}\) \\ \hline
GHT-PS                 & \multicolumn{1}{c|}{0.5/0.7/O} & \multicolumn{1}{c|}{0.3/0.3/\(\*A\)} & \multicolumn{1}{c|}{0.3/0.7/\(\*O\)} & \multicolumn{1}{c|}{0.3/1/\(\hat{\*V}\)}   & 0.5/0.3/\(\*A\) & \multicolumn{1}{c|}{0.5/0.5/\(\hat{\*V}\)} & 0.3/0.3/\(\hat{\*V}\) \\ \hline
\end{tabular}
\caption{The configuration of \(\alpha\), \(\beta\), and Feature Maps (FM, including \(\hat{\*V}\), \(\*A\), and \(\*O\)) for GHT and GHT-PS on different Machine Translation datasets.}
\label{tab:MTconfig}
\end{table*}

\begin{table}[H]
\centering
\begin{tabular}{|c|c|}
\hline
Model  & \(\alpha/\beta/FM\) \\ \hline
GHT    & 0.5/0.5/\(\hat{\*V}\)     \\ \hline
GHT-PS & 0.5/0.5/\(\hat{\*V}\)     \\ \hline
\end{tabular}
\caption{The configuration of \(\alpha\), \(\beta\), and Feature Maps (FM, including \(\hat{\*V}\), \(\*A\), and \(\*O\)) for GHT and GHT-PS in Abstractive Summarization and Language Modeling.}
\label{tab:LMASconfig}
\end{table}

\subsubsection{Language Modeling}
We evaluate LM on WIKITEXT-103~\citep{DBLP:conf/iclr/MerityX0S17} which has about 100M tokens and a 260K BPE vocabulary.
Following~\citet{DBLP:conf/iclr/BaevskiA19}, we use perplexity as an evaluation metric and a context window of 2047 at the inference stage.

\subsubsection{Abstractive Summarization} 
We also evaluate on CNN-DailyMail~\citep{DBLP:conf/nips/HermannKGEKSB15} for AS to test the ability of GHT in hard tasks with long inputs. The dataset comprises over 280K news articles paired with multi-sentence summaries.
Following~\citet{DBLP:conf/iclr/WuFBDA19}, articles are truncated to 512 tokens and encoded with 50K BPE. We use F1-Rouge~\citep{lin-2004-rouge} to evaluate the performance, including Rouge-1, Rouge-2 and Rouge-L.


\end{document}